\documentclass[10pt,twocolumn,letterpaper]{article}
\usepackage{cvpr}              

\usepackage{graphicx}
\usepackage{amsmath}
\usepackage{amssymb}
\usepackage{booktabs}

\usepackage{times}
\usepackage{epsfig}

\usepackage{verbatim}
\usepackage{helvet}
\usepackage{courier}
\usepackage{amsfonts}
\usepackage{caption}
\graphicspath{ {./images/} }
\usepackage{threeparttable}
\usepackage{multicol}
\usepackage{multirow}
\usepackage{placeins}
\usepackage[accsupp]{axessibility} 

\usepackage[pagebackref=true,breaklinks=true,letterpaper=ture,colorlinks,bookmarks=false]{hyperref}

\usepackage[capitalize]{cleveref}
\crefname{section}{Sec.}{Secs.}
\Crefname{section}{Section}{Sections}
\Crefname{table}{Table}{Tables}
\crefname{table}{Tab.}{Tabs.}

\usepackage{tabularx}
\newcolumntype{L}[1]{>{\raggedright\arraybackslash}p{#1}}
\newcolumntype{C}[1]{>{\centering\arraybackslash}p{#1}}
\newcolumntype{R}[1]{>{\raggedleft\arraybackslash}p{#1}}

\def\cvprPaperID{3056} 
\def\confName{CVPR}
\def\confYear{2022}

\begin{document}

\title{FS6D: Few-Shot 6D Pose Estimation of Novel Objects}

\author{Yisheng He$^{1}$ \quad {Yao Wang}$^{2}$ \quad {Haoqiang Fan}$^{2}$ \quad {Jian Sun}$^{2}$ \quad {Qifeng Chen}$^{1}$ \\
	${^1}$Hong Kong University of Science and Technology \quad
	${^2}$Megvii Technology \\
}
\maketitle

\begin{abstract}
6D object pose estimation networks are limited in their capability to scale to large numbers of object instances due to the close-set assumption and their reliance on high-fidelity object CAD models. In this work, we study a new open set problem; the few-shot 6D object poses estimation: estimating the 6D pose of an unknown object by a few support views without extra training. To tackle the problem, we point out the importance of fully exploring the appearance and geometric relationship between the given support views and query scene patches and propose a dense prototypes matching framework by extracting and matching dense RGBD prototypes with transformers. Moreover, we show that the priors from diverse appearances and shapes are crucial to the generalization capability under the problem setting and thus propose a large-scale RGBD photorealistic dataset (ShapeNet6D) for network pre-training. A simple and effective online texture blending approach is also introduced to eliminate the domain gap from the synthesis dataset, which enriches appearance diversity at a low cost. Finally, we discuss possible solutions to this problem and establish benchmarks on popular datasets to facilitate future research. [\href{https://fs6d.github.io/}{project page}]
\end{abstract}
\begin{figure}
    \centering
    \includegraphics[scale=0.65]{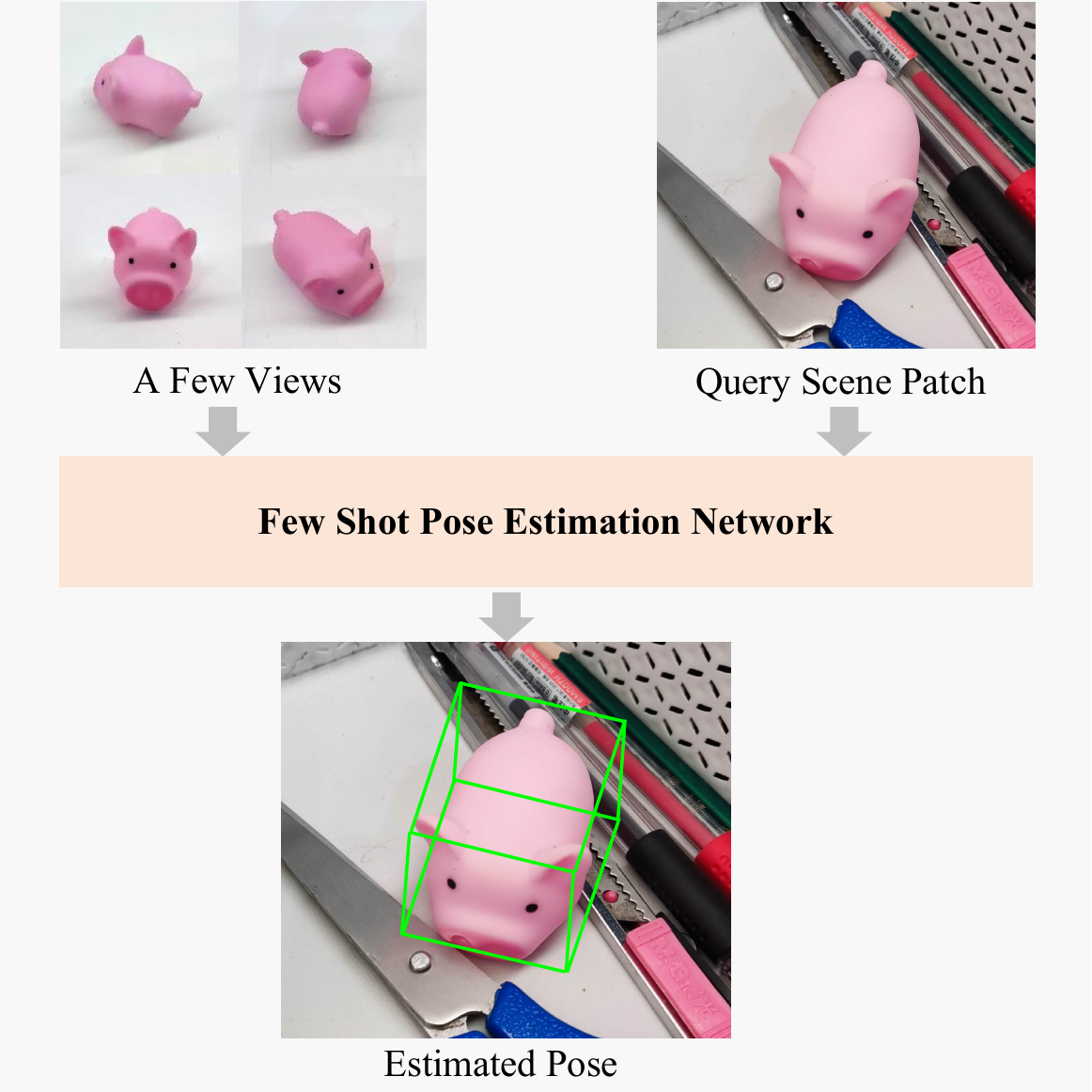}
    \caption{
        \textbf{The few-shot 6D pose estimation problem}. Given a few RGBD views of a \textit{novel} objects with pose labels. The few-shot pose estimation network aims to estimate 6D pose of that object in a novel query scene without extra training. No precise CAD models are required as well.
    }
    \label{fig:intro}
\end{figure}

\section{Introduction}
6D object pose estimation aims to predict a rigid transformation from the object coordinate system to the camera coordinate system, which benefits various applications, including robotic manipulation, augmented reality, autonomous driving, etc. The explosive development of deep learning has brought significant improvement to this problem. With recent works \cite{he2020pvn3d,FFB6D} reaching nearly 99\% recall accuracy on existing benchmarks \cite{hinterstoisser2011multimodal,xiang2017posecnn,bop2020}, one may get the impression that the 6D object pose problem has been solved, which is not the case. We argue that the current problem has been simplified with strict restrictions. They are under the \textit{close-set} assumption that the training and testing data are drawn from the same object space, which, however, does not adhere to the real dynamic worlds. Moreover, extravagant high-fidelity CAD models and large-scale datasets are required for training to obtain good performance on new objects under the current instance-level pose estimation setting. 

The recently proposed category-level pose estimation task\cite{wang2019normalized} loosens the restriction with generalizability to novel objects within the same categories. However, it is still limited in the \textit{close-set} assumption of predefined categories. 
Instead, in this work, we study a new \textit{open-set} problem, the few-shot 6D object pose estimation: estimating 6D pose of unknown objects by only a few views of the objects without extra training. As shown in Figure \ref{fig:intro}, in our setting, only a few labeled RGBD images of novel objects are provided, and no high-fidelity CAD models are required. The goal of the problem is to bridge the capability gap between machine learning algorithms and flexible human visual systems that can locate and estimate the pose of a novel object given only several views of it. Besides, it has a wide range of real-world applications in robotic vision systems, i.e., fast registration of novel objects for robotic manipulation and home robots.

Under the observation that human beings utilize both appearance and geometric information to match and locate a new object, we propose a dense RGBD prototypes matching framework to tackle the problem. Specifically, transformers are utilized to fully explore the semantic and geometric relationship between the query scene patch and the support views of novel objects. Moreover, we point out that large-scale datasets' diverse shape and appearance priors are essential to empower networks to generalize on novel objects. Therefore, we introduce a large-scale photorealistic dataset (ShapeNet6D) with diverse shapes and appearances for prior learning. To our knowledge, ours (800K images of 12K objects) is the largest and most diverse dataset for 6D pose algorithms. To bridge the domain gap between rendered RGB images and real-world scenes, we introduce a simple and effective \textit{online} texture blending augmentation, which further enriches the appearance diversity and facilitates network performance at a low cost.   

To summarize, the contributions of this work are:
\begin{itemize}
    \item We introduce a challenging \textit{open-set} problem, the few-shot 6D object pose estimation, and establish a benchmark to study it.
    \item We formulate the problem by dense RGBD prototypes matching and introduce FS6D-DPM, which fully leverage appearance and geometric information to tackle the problem.
    \item Datasets: We introduce ShapeNet6D, a large-scale photorealistic dataset with diverse shapes and appearances for prior learning of few-shot 6D pose estimation algorithms. We also introduce an online texture blending augmentation to obtain scenes of texture-rich objects without domain gaps at a low cost. 
\end{itemize}

\section{Related Work}
\subsection{6D Object Pose Estimation in Close-Set Setting}
Instance-level pose estimation retrieves pose parameters of \textit{known} object instances. Matching based approaches \cite{huttenlocher1993comparing,gu2010discriminative,hinterstoisser2011gradient,sundermeyer2020multi,xiao2019pose} requires precise CAD models to render thousands of templates and establish hand-craft or learned codebook for matching. Learning-based approaches includes direct pose regression \cite{xiang2017posecnn,wang2019densefusion}, dense correspondence exploration \cite{li2019cdpn} and recent keypoint-based approaches \cite{he2020pvn3d,FFB6D,peng2019pvnet}, which improve the performance by large margins. Despite compelling results, these approaches can only deal with scenarios of \textit{known objects} with high-fidelity CAD models. Instead, the recent category-level pose estimation \cite{wang2019normalized} improves the generalizability by estimating unseen object instances within the \textit{known categories}. Normalized Object Coordinate Space (NOCS) \cite{wang2019normalized} or shape deformation based \cite{he2022towards,tian2020SPD} approaches are proposed. However, both traditional instance- and category-level pose estimation problems are under the \textit{close-set} setting, assuming that the training and testing data are within the same predefined instance or category spaces. While such \textit{close-set} setting does not adhere to the real dynamic world, we instead define a new \textit{open-set} problem, the few-shot 6D pose estimation. Algorithms developed in our \textit{open-set} setting can be flexibly applied to unknown objects without extra training with only a few labeled RGBD images, no matter they are within the trained categories or not. 

\subsection{Possible Few-Shot Pose Estimation Solutions}
\textbf{Local Image Feature Matching.} Local feature matching can establish the correspondence between two images for the few-shot pose estimation problem. Existing methods can be categorized into detector-based \cite{sarlin2020superglue,rublee2011orb,lowe1999object,luo2018geodesc,luo2020aslfeat} and detector-free \cite{sun2021loftr,li2020dual,liu2010sift,rocco2018neighbourhood}. While these algorithms only leverage the grey-scale images, the performance drops on texture-less objects. Instead, we fully leverage both the appearance and the geometric information and generalize well in more scenarios.

\textbf{Point Cloud Registration.} One line of point cloud registration algorithms solve the problem by detecting 3D keypoints \cite{bai2020d3feat,li2019usip}, extracting feature descriptors \cite{gofutherPPF,gojcic2019perfect,choy2019fully,poiesi2021distinctive,poiesi2021distinctive,deng2018ppfnet} and estimating the relative transformation. Several end-to-end approaches \cite{wang2019prnet} are also proposed. However, these algorithms heavily rely on fine point clouds and fail on objects that are not captured by depth sensors, i.e., reflective ones. Instead, we fully leverage the complementary information in RGBD images for dense prototypes extraction and matching to retrieve better object pose parameters.

\subsection{Metric learning on few-shot learning problems} 
Metric learning techniques have been applied to several few-shot learning problems, including classification \cite{garcia2017few,vinyals2016matching,snell2017prototypical} and segmentation \cite{dong2018few,tian2020prior,yang2020prototype,liu2020part}. The representative prototypical network \cite{snell2017prototypical} for classification map the support and query images into a global embedding space and then retrieve the class label of query image based on the support embedding, named prototype. The recent metric learning-based approaches in more challenging segmentation areas utilize similar technique but output per-pixel prediction on the query images by matching per-pixel query features with global average prototypes \cite{dong2018few,zhang2019canet,tian2020prior} or part-level prototypes \cite{yang2020prototype,liu2020part}. While sparse support prototypes are enough to solve the above problems, few-shot pose estimation requires more dense correspondence exploration on pixel-level support prototypes and query features, which is more challenging.
\section{Proposed Method}

\subsection{Problem Formulation}
\label{sec:pf}
We introduce the problem setting of the few-shot 6D object pose estimation and the derived domain generalization problem.

\textbf{The few-shot 6D object pose estimation.} We formulate the new \textit{open-set} task, the few-shot 6D pose estimation as follows. Given $k$ support RGBD patches $P=\{p_1, p_2, ..., p_k\}$ of a novel object with pose parameters as support frames, the inference task is to retrieve the 6D pose parameters of that novel object in the query novel scene image $I$. Compared to current \textit{close-set} setting, the proposed \textit{open-set} one eliminates the reliance on precise CAD models and focuses on the generalizability of trained models on unseen objects. Specifically, once the model is trained, we expect to apply it on novel scenes of novel objects by a few views without extra training. It bridges the gap between machine learning algorithms and flexible human visual systems. Moreover, it enables real-world applications, i.e., fast registration of new objects for robotic manipulation and service home robots. 

The generalization requirement of the \textit{open-set} problem also derive another interesting research question to study:

\textbf{Domain generalization.} The domain generalization targets to reduce the domain gaps between models trained on the synthesis and real-world data. It has been introduced to the 6D pose estimation field to deal with the lack of data \cite{su2015render,kehl2017ssd,manhardt2018deep,wang2020self6d}. However, this field is less explored as existing real-world benchmarking datasets \cite{hodan2017tless,bop2020} for the \textit{close-set} problem has been well established: real-world training data for the object to be estimated is available. While existing datasets are small with limited objects, in our few-shot \textit{open-set} setting, diversity of shape and appearance are crucial to the generalizability of few-shot 6D object pose estimation algorithms. However, capturing and labeling such a large-scale real-world dataset is not practical due to the high cost (money and time). It is crucial to fully leverage the geometry and appearance diversity in our large-scale photorealistic datasets and generalize to the real world. The domain generalization problem is thus an important problem to study for the few-shot 6D pose estimation.

\begin{figure*}
    \centering
    \includegraphics[scale=0.4]{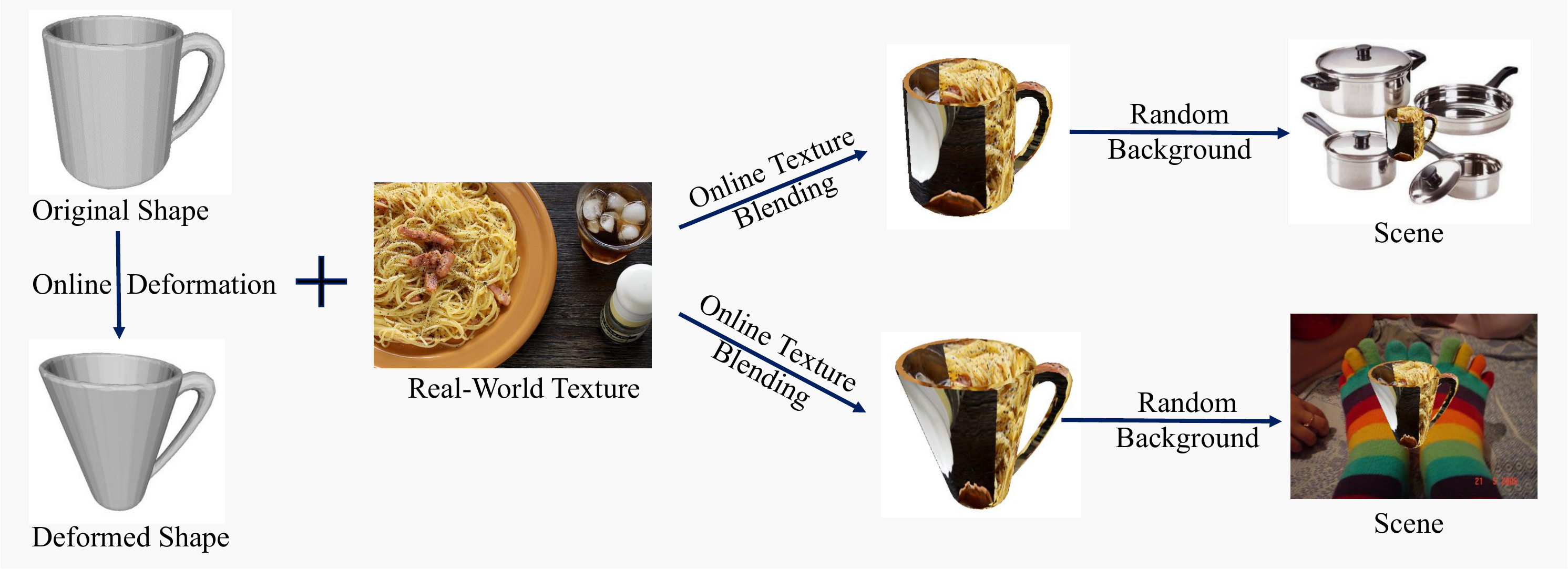}
    \caption{
        \textbf{Online data augmentation.} The \textit{online} texture blending augmentation generates texture by directly blending the real-world image to the object mesh model. No extra artificial simulation is applied, i.e., simulated lighting and the domain of real-world RGB images is preserved. Along with the online deformation augmentation \cite{chen2021fsnet}, we can obtain data with diverse appearances and shapes at a low cost.
    }
    \label{fig:online_texture_blending}
\end{figure*}

\subsection{Datasets} 
\label{sec:sn6d}

The prior learned from large-scale datasets is crucial to the performance and generalizability of few-shot learning algorithms. ImageNet \cite{deng2009imagenet}, for example, has been widely used for network pre-training in several few-shot learning tasks, i.e., object detection and segmentation. While 2D vision tasks rely more on the semantic prior in RGB images, for the few-shot 6D object pose estimation, both shape and semantic prior are crucial for the generalizability of the network. However, existing datasets \cite{xiang2017posecnn,hodan2017tless,bop2020} for 6D object pose estimation are small and lack diversity in shape and appearance to provide enough prior for the generalization capability. Therefore, we keep their role as real-world benchmark datasets and propose a new large-scale dataset, ShapeNet6D, with diverse shapes and appearances for prior learning.

\renewcommand{\arraystretch}{1.3}
\begin{table}[tp]
    \centering
    \fontsize{8.}{8.}\selectfont
     \begin{tabular}{l|l|llllllll}
     \hline
    Dataset    & Modality & $N_{cat}$ & $N_{obj}$ & $N_{img}$      \cr\hline  
    LineMOD \cite{hinterstoisser2011multimodal}   & RGBD     & -    & 15      & 18,273    \\ 
    YCB-V \cite{calli2015ycb}    & RGBD     & -    & 21      & 133,936   \\ 
    TLESS \cite{hodan2017tless}     & RGBD     & -    & 30       & 47,664    \\
    NOCS-REAL \cite{wang2019normalized} & RGBD     & 6    & 42       &  80,000      \\
    NOCS-CAMERA \cite{wang2019normalized} &RGBD     & 6    & 1,085     &  300,000      \\
    ShapeNet6D & RGBD     & \textbf{51}     & \textbf{12,490}  & \textbf{800,000} \cr\hline 
    \end{tabular}
    \caption{\textbf{Statistics of Different Datasets.} ShapeNet6D is diverse in shape and appearance, which is crucial to the generalizability of few-shot 6D pose algorithms. $N_{cat}$: number of category; $N_{obj}$: number of object instance. $N_{img}$: number of images.}
    \label{tab:ds_cmp}
\end{table}

\subsubsection{ShapeNet6D} The proposed ShapeNet6D is a large-scale photorealistic dataset containing RGBD scene images of more than 12K object instances from the ShapeNet \cite{chang2015shapenet} repository. Each scene image is labeled with ground truth information for the 6D pose estimation problem, including instance semantic segmentation and pose parameters of each object. As we demonstrate empirically, the diversity of shape and appearance is crucial for the network to generalize. While it is not practical to collect and label such a large-scale, diverse dataset in the real world due to the high cost (time and money), we instead generate photorealistic images by physically-based rendering. Our approach is inspired by the successful application of photorealistic datasets in \cite{yao2020blendedmvs,bop2020,yang2021ishape} while improving the diversity of object shape and appearance. Specifically, we utilize the physically-based rendering engine, Blender\footnote{https://www.blender.org} that simulates the flow of light energy by ray tracing to render realistic scene images. To arrange a scene to render, we first randomly select several objects from ShapeNet, apply random material and texture, and drop them into a box with the PyBullet physics engine integrated into Blender. To enrich the variety of the background, we randomly selected physically-based rendering material from the HDRI Haven\footnote{https://hdrihaven.com/hdris} and applied them to the wall of the box. Random environment lights are also added to generate diverse lighting conditions. Finally, the RGBD scene image is rendered from a random camera pose, and the ground truth instance semantic segmentation labels and pose parameters of each object are also obtained. Statistics about ShapeNet6D compared to existing 6D pose benchmark datasets are shown in Table \ref{tab:ds_cmp}. ShapeNet6D is on a larger scale and is more diverse in shape and appearance, which provides better prior to the few-shot pose estimation problem as we showed empirically. 

\subsubsection{Online texture blending}
As one of the crucial clues to solve the few-shot 6D pose estimation problem, the texture field is also essential to the performance of the few-shot 6D object pose estimation. However, it is labor-intensive and time-consuming to generate textures and materials for objects that can be rendered to be photorealistic. The rendered RGB images tend to have more significant domain gaps between the real world as well. Moreover, to produce photorealistic images, time- and computation-consuming techniques like ray tracing are required. Therefore, the images should be pre-processed offline and stored before network training, which costs a lot of storage space for a large-scale dataset.
On the other hand, real-world RGB images captured from various cameras are easy to access, i.e., ImageNet \cite{deng2009imagenet}, and MS-COCO \cite{lin2014MS_COCO}. It motivates us to leverage efficient texture wrapping techniques to generate scenes of objects with rich real-world texture to serve as \textit{online} data argumentation. Specifically, the mesh is first unwrapped to obtain a UV map. For each triangle, we get the UV coordinate of each vertex and then utilize it to determine the UV coordinate of each pixel by linear interpolation during rasterization. The UV coordinate is then applied to lookup the color value from a texture map randomly sampled from the real-world ImageNet \cite{deng2009imagenet}, and MS-COCO \cite{lin2014MS_COCO}. Previous works \cite{bop2020,park2020latentfusion} render images with artificial simulations, i.e., Beckmann model \cite{beckmann1987scattering} , which change the domain and cause domain gaps. Instead, we applied no simulation, so the composite images are kept in the real domain, i.e., the lighting condition, sensors noise of the real-world images are preserved. Moreover, such a simple blending strategy can be implemented fast to serve \textit{online}. Moreover, we can combine it with \textit{online} shape deformation \cite{chen2021fsnet} to produce data with rich appearance and shape diversity for training, as shown in Figure \ref{fig:online_texture_blending}.



\begin{figure*}
    \centering
    \includegraphics[scale=0.54]{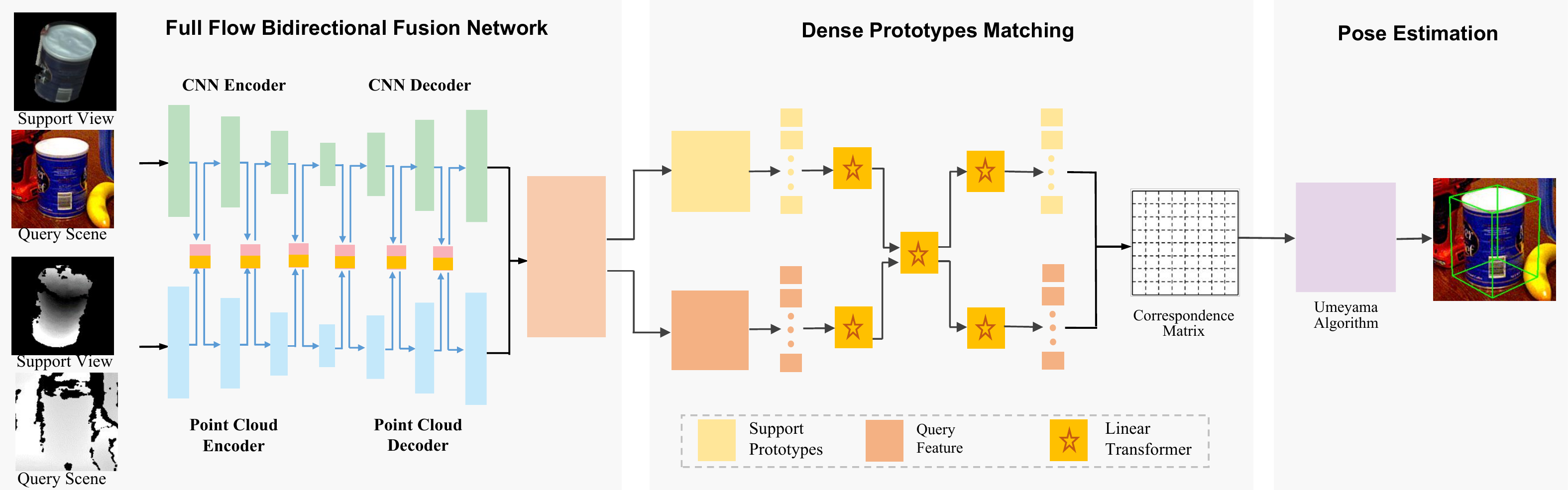}
    \caption{
        \textbf{Overview of our pipeline.} A Siamese full flow bidirectional fusion network \cite{FFB6D} is utilized to extract rich appearance and geometric features from the support view and the query scene patch, respectively. The extracted features are then fed into self- and cross-attention modules to obtain dense support prototypes and query features for correspondence reasoning. Finally, the Umeyama algorithm \cite{umeyama1991least} is applied to recover the pose parameters of novel objects in the query scene patch.
    }
    \label{fig:network}
\end{figure*}

\subsection{FS6D-DPM}
\label{sec:dpm}
\subsubsection{Preliminaries}
\textbf{Prototypes-based few-shot learning.} We first briefly introduce the prototypes-based algorithms for few-shot learning. It has been successfully applied to various few-shot 2D vision tasks, i.e., classification and semantic segmentation. Specifically, a pre-trained Siamese backbone is utilized for feature extraction from the support and the query images. Then, global average pooling is applied on the extracted support feature maps to obtain the support prototypes. 
This global average prototype is then applied to calculate the similarity between the global features (in classification) or dense pixel-wise features (in semantic segmentation) extracted from the query image for prediction. However, these tasks' global-to-global or global-to-local correspondence is not enough to recover 6D object pose parameters. This work, instead, proposes a dense prototypes extraction module to establish the local-to-local correspondence between the support RGBD images and the query scene patch for pose estimation. 

\textbf{Transformer \cite{vaswani2017attention}.}
Transformers networks are first introduced in Natural Language Processing and are brought into many vision tasks. The multi-head attention mechanism enables it to capture the long-term dependency even on an unordered set. Specifically, given three vectors as inputs, namely query $Q$, key $K$, and value $V$. The attention mechanism is to retrieve information $I$ from the value s.t. the similarity between $Q$ and $K$, denoted as:
\begin{equation}
    I_{retrieved} = softmax(QK^T)V. 
\end{equation}

Gifted with the capability of capturing long-term dependency, the Transformer networks \cite{vaswani2017attention,wang2020linear_former} have been successfully applied to aggregate contextual information in the local feature matching \cite{sarlin2020superglue,sun2021loftr} and point cloud registration \cite{huang2021predator} field. In this work, we further extend it to dense RGBD prototypes matching for few-shot 6D pose estimation.


\begin{figure}
    \centering
    \includegraphics[scale=0.5]{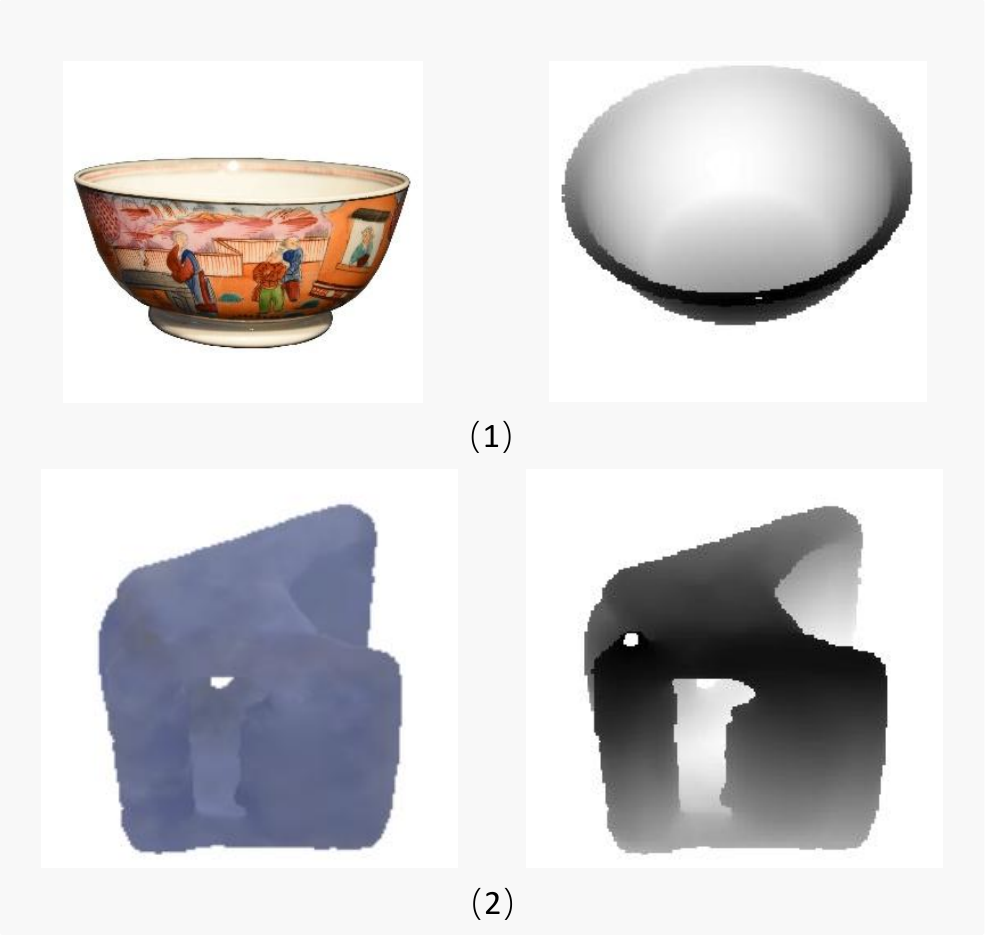}
    \caption{
    \textbf{Complementary information in RGBD images for few-shot 6D pose estimation.} (1) Texture information in RGB images is crucial cue for objects with smooth surface. (2) Geometric information in depth images is important cue for texture-less objects. 
    }
    \label{fig:texture_depth}
\end{figure}

\subsubsection{Overview}
To build a few-shot pose estimation algorithm that can generalize well to novel objects, it is crucial to fully explore the semantic and geometric relationship between the given support views and the query scene patch, as shown in Figure \ref{fig:texture_depth}. In this section, we introduce our dense prototypes matching framework to tackle this challenging problem. As shown in Figure \ref{fig:network}, our framework consists of three main parts. Firstly, a Siamese RGBD feature extraction backbone is utilized to extract rich semantic and geometric features for each pixel/point. Then, a dense prototypes extraction network based on transformers is applied to extract dense RGBD prototypes from the support view and point-wise local features from the query scene patch for similarity calculation. Finally, after the correspondence between dense prototypes and scene features is established, the Umeyama algorithm \cite{umeyama1991least} is leveraged to estimate the 6D pose parameters.

\subsubsection{Feature Extraction Backbone}
The first step is to extract rich semantic and geometric features from the given RGBD images. As a fundamental problem, many works \cite{FFB6D,wang2019densefusion,xu2018pointfusion} have studied this representation learning task. Recently, FFB6D \cite{FFB6D} introduce a full flow bidirectional fusion network for 6D pose estimation and significantly improve the performance of \textit{close-set} pose estimation. Specifically, bidirectional local feature fusion blocks are added into each encoding and decoding layer to bridge the information gap and improve the quality of extracted semantic and geometric features (see \cite{FFB6D} for details). In this work, we leverage FFB6D to build a Siamese network for feature extraction from the support images and the query scenes.

\subsubsection{Dense Prototypes Extraction and Matching}

Now we have obtained dense features from the Siamese feature extraction backbone. We then extract dense support prototypes and query features to calculate the similarity and establish the correspondence. To extract descriptive and representative dense RGBD prototypes from the support views and dense query features from query scenes, it is crucial to fully leverage the structural geometric information residing in point clouds and semantic information abiding in RGB images. Besides, contextual information between the support shot and the query patch is also essential to improve the precision of similarity calculation and correspondence exploration. 

Considering the power of transformers on long-term dependency capturing, we utilize the optimized Linear Transformers \cite{wang2020linear_former} to serve the above two purposes. As shown in the middle part of Figure \ref{fig:network}, we first establish self-attention on the extracted feature maps to strengthen the geometric and semantic information residing in the extracted dense prototypes and dense query features. We regard the extracted features as query, key, and value and fed them into the Linear Transformer networks to enhance the semantic and geometric features. Meanwhile, a cross-attention module is also applied to explore the contextual information between the support prototypes and the query scene features. Precisely, to extract contextual information from the support prototypes to the query scene features, we took each scene feature as a query and the dense prototypes as keys and values to the Linear Transformers. Contextual information from query scene features to support prototypes is enhanced similarly.  With extracted contextual information, another self-attention modules are applied to enhance the geometric and semantic features further. In this way, we obtain dense support prototypes and query features with rich semantic, geometric and contextual information. Unlike prototype-based few-shot classification and segmentation algorithms that calculate the similarity by cosine distance, we follow local feature matching pipelines \cite{sarlin2020superglue} to establish the dense correspondence by calculating $C(i, j) = \langle P(i), Q(j)\rangle$ with $P(i)$ the $i_{th}$ prototype, $Q(j)$ the $j_{th}$ query feature and $\langle\cdot,\cdot\rangle$ the inner product. The Sinkhorn Algorithm \cite{peyre2019computational} is applied for differentiable optimization as well.

\subsubsection{Pose Parameters Estimation}
After the correspondence between the dense prototypes and the query scene features is established, we utilize the Umeyama \cite{umeyama1991least} algorithms to recover the pose parameters. Specifically, given a set of matched pairs $\mathcal{M} = \{(p_i, q_i), 1 \leq i \leq N\}$ with $p_i$, $q_i$ the 3D coordinate of matched prototypes and queries, the Umeyama algorithms estimate the rotation $R$ and translation $T$ by minimizing:
\begin{equation}
    L_{lsq} = \sum_{i=1}^N||q_i - (Rp_i+T)||_2^2.
\end{equation}
To eliminate the influence of outliers. The RANSAC algorithms are also applied.

Given $K$ support views of a novel object, we can obtain $K$ predicted pose parameters along with their losses. We select the one with minimum loss as our final prediction.

\begin{figure*}
  \centering
     \includegraphics[scale=0.52]{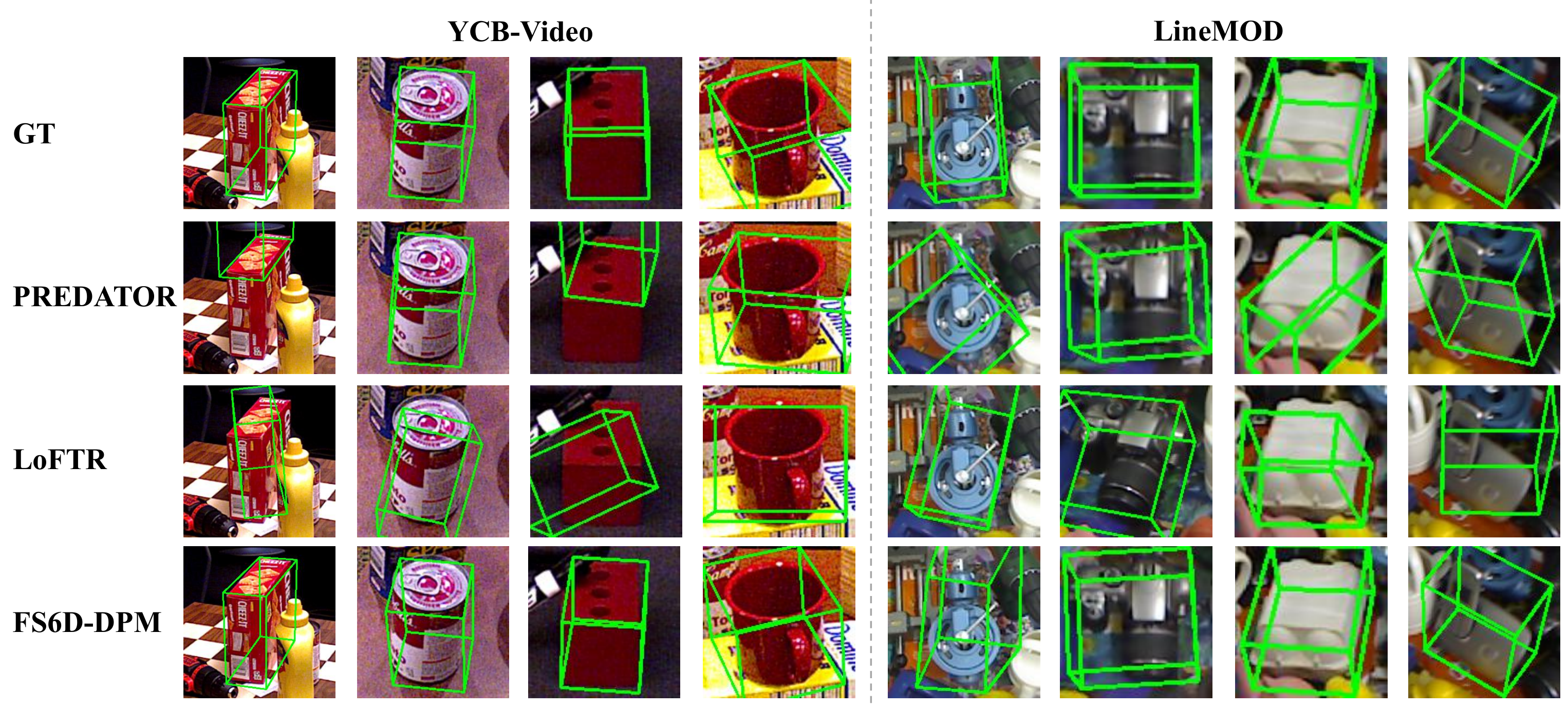}
     \caption{
        Qualitative results on the YCB-Video (left) and the LineMOD (right) datasets. We visualize the results of PREDATOR \cite{huang2021predator}, LoFTR \cite{sun2021loftr} and the proposed FS6D-DPM. The ground truths are also visualized in the first row.
     }
     \label{fig:Qualitative_Results}
\end{figure*}

\section{Experiments}

\begin{table*}[]
    \centering
    \newcommand{\ycbC}{1.1}
    \fontsize{8.}{8.}\selectfont
    \begin{tabular}{C{\ycbC cm}C{3.3 cm}|C{\ycbC cm}C{\ycbC cm}|C{\ycbC cm}C{\ycbC cm}|C{\ycbC cm}C{\ycbC cm}|C{\ycbC cm}C{\ycbC cm}}
    \hline
    \multicolumn{1}{c|}{\multirow{2}{*}{Group}} & \multirow{2}{*}{Object} & \multicolumn{2}{c|}{PREDATOR \cite{huang2021predator}} & \multicolumn{2}{c|}{LoFTR \cite{sun2021loftr}} & \multicolumn{2}{c|}{TP-UB} & \multicolumn{2}{c}{FS6D-DPM} \\ \cline{3-10} 
    \multicolumn{1}{c|}{}                       &                         & ADDS          & ADD        & ADDS        & ADD       & ADDS           & ADD          & ADDS         & ADD        \\ \hline
\multicolumn{1}{c|}{\multirow{7}{*}{0}} & 002 master chef can     & 73.0                     & 17.4                       & 87.2                     & \textbf{50.6}              & 62.2                     & 21.4                       & \textbf{92.6}            & 36.8                       \\
\multicolumn{1}{c|}{}                   & 003 cracker box         & 41.7                     & 8.3                        & 71.8                     & \textbf{25.5}              & 65.6                     & 5.0                        & \textbf{83.9}            & 24.5                       \\
\multicolumn{1}{c|}{}                   & 004 sugar box           & 53.7                     & 15.3                       & 63.9                     & 13.4                       & 66.7                     & 21.5                       & \textbf{95.1}            & \textbf{43.9}              \\
\multicolumn{1}{c|}{}                   & 005 tomato soup can     & 81.2                     & 44.4                       & 77.1                     & 52.9                       & 75.2                     & 43.1                       & \textbf{93.0}            & \textbf{54.2}              \\
\multicolumn{1}{c|}{}                   & 006 mustard bottle      & 35.5                     & 5.0                        & 84.5                     & 59.0                       & 47.1                     & 4.0                        & \textbf{97.0}            & \textbf{71.1}              \\
\multicolumn{1}{c|}{}                   & 007 tuna fish can       & 78.2                     & 34.2                       & 72.6                     & \textbf{55.7}              & 72.8                     & 38.4                       & \textbf{94.5}            & 53.9                       \\
\multicolumn{1}{c|}{}                   & 008 pudding box         & 73.5                     & 24.2                       & 86.5                     & 68.1                       & 86.3                     & 18.4                       & \textbf{94.9}            & \textbf{79.6}              \\ \hline
\multicolumn{1}{c|}{\multirow{7}{*}{1}} & 009 gelatin box         & 81.4                     & 37.5                       & 71.6                     & \textbf{45.2}              & 90.9                     & 43.2                       & \textbf{98.3}            & 32.1                       \\
\multicolumn{1}{c|}{}                   & 010 potted meat can     & 62.0                     & 20.9                       & 67.4                     & 45.1                       & 59.8                     & 28.9                       & \textbf{87.6}            & \textbf{54.9}              \\
\multicolumn{1}{c|}{}                   & 011 banana              & 57.7                     & 9.9                        & 24.2                     & 1.6                        & 79.2                     & 54.5                       & \textbf{94.0}            & \textbf{69.1}              \\
\multicolumn{1}{c|}{}                   & 019 pitcher base        & 83.7                     & 18.1                       & 58.7                     & 22.3                       & 17.5                     & 0.7                        & \textbf{91.1}            & \textbf{40.4}              \\
\multicolumn{1}{c|}{}                   & 021 bleach cleanser     & 88.3                     & \textbf{48.1}              & 36.9                     & 16.7                       & 20.3                     & 0.6                        & \textbf{89.4}            & 44.1                       \\
\multicolumn{1}{c|}{}                   & 024 bowl                & 73.2                     & \textbf{17.4}              & 32.7                     & 1.4                        & 30.7                     & 0.0                        & \textbf{74.7}            & 0.9                        \\
\multicolumn{1}{c|}{}                   & 025 mug                 & 84.8                     & 29.5                       & 47.3                     & 23.6                       & 46.0                     & 13.9                       & \textbf{86.5}            & \textbf{39.2}              \\ \hline
\multicolumn{1}{c|}{\multirow{7}{*}{2}} & 035 power drill         & 60.6                     & 12.3                       & 18.8                     & 1.3                        & 42.3                     & 0.7                        & \textbf{73.0}            & \textbf{19.8}              \\
\multicolumn{1}{c|}{}                   & 036 wood block          & 70.5                     & 10.0                       & 49.9                     & 1.4                        & 13.5                     & 1.3                        & \textbf{94.7}            & \textbf{27.9}              \\
\multicolumn{1}{c|}{}                   & 037 scissors            & 75.5                     & 25.0                       & 32.3                     & 14.6                       & \textbf{89.5}            & \textbf{71.8}              & 74.2                     & 27.7                       \\
\multicolumn{1}{c|}{}                   & 040 large marker        & 81.8                     & 38.9                       & 20.7                     & 8.4                        & 82.5                     & 51.9                       & \textbf{97.4}            & \textbf{74.2}              \\
\multicolumn{1}{c|}{}                   & 051 large clamp         & \textbf{83.0}            & 34.4                       & 24.1                     & 11.2                       & 49.0                     & 20.0                       & 82.7                     & \textbf{34.7}              \\
\multicolumn{1}{c|}{}                   & 052 extra large clamp   & \textbf{72.9}            & \textbf{24.1}              & 15.0                     & 1.8                        & 50.2                     & 9.4                        & 65.7                     & 10.1                       \\
\multicolumn{1}{c|}{}                   & 061 foam brick          & 79.2                     & 35.5                       & 59.4                     & 31.4                       & 91.8                     & \textbf{60.5}              & \textbf{95.7}            & 45.8                       \\ \hline
\multicolumn{2}{c|}{MEAN}                                           & 71.0                     & 24.3                       & 52.5                     & 26.2                       & 59.0                     & 24.2                       & \textbf{88.4}            & \textbf{42.1}  \\ \hline
    \end{tabular}
        \caption{Quantitative evaluation of different few-shot 6D pose baselines on the YCB-Video dataset. Among them, the proposed FS6D-DPM fully leverages the appearance and geometric information achieves the best performance. TP-UB: upper bound of template approaches. }
        \label{tab:ycb_result}
    \end{table*}

\begin{table*}[]
    \centering
    \newcommand{\linemodC}{2.3}
    \fontsize{8.0}{8.0}\selectfont
    \begin{tabular}{C{\linemodC cm}|C{\linemodC cm}|C{\linemodC cm}|C{\linemodC cm}|C{\linemodC cm}}
    \hline
          \multirow{2}{*}{Group} & PREDATOR \cite{huang2021predator} & LoFTR \cite{sun2021loftr} & TP-UB & FS6D-DPM     \\ \cline{2-5} 
                       & ADD-0.1d & ADD-0.1d & ADD-0.1d    & ADD-0.1d \\ \hline
    0     & 55.1     & 38.0     & 8.1         & \textbf{70.0}     \\
    1     & 40.4     & 30.4     & 10.0        & \textbf{86.8}     \\
    2     & 46.8     & 30.3     & 13.2        & \textbf{93.4}     \\ \hline
    Mean  & 48.0     & 33.4     & 10.1        & \textbf{83.4}     \\ \hline
    \end{tabular}
    \caption{Quantitative evaluation of different few-shot 6D pose baselines on the LineMOD dataset. The proposed FS6D-DPM that fully leverages the appearance and geometric information achieves the best performance. TP-UB: upper bound of template-based approach.}
    \label{tab:linemod_result}
\end{table*}

\begin{table}[]
    \newcommand{\ycbC}{1.5}
    \centering
    \fontsize{8.0}{8.0}\selectfont
    \begin{tabular}{l|C{\ycbC cm}|C{\ycbC cm}}
    \hline
    \multirow{2}{*}{Object} & \multicolumn{1}{c|}{w/o OTB} & \multicolumn{1}{c}{w/ OTB} \\ \cline{2-3} 
                            & ADD                    & ADD                     \\ \hline
    002 master chef can     & 23.4                     & \textbf{50.0}                     \\
    003 cracker box         & 15.1                     & \textbf{42.0}                     \\
    004 sugar box           & 12.3                     & \textbf{52.5}                     \\
    005 tomato soup can     & 52.8                     & \textbf{74.7}                     \\
    006 mustard bottle      & 55.4                     & \textbf{75.4}                    \\
    007 tuna fish can       & 54.5                     & \textbf{56.5}                     \\
    008 pudding box         & 34.4                     & \textbf{42.2}                     \\
    009 gelatin box         & 50.7                     & \textbf{94.2}                     \\
    010 potted meat can     & 38.7                      & \textbf{54.8}        \\
    \hline
    Mean                    & 37.5                      & \textbf{60.3}        \\
    \hline
    \end{tabular}
    \caption{Effect of online texture blending. w/o OTB: without online texture blending; w/ OTB: with online texture blending.}
    \label{tab:online_wrapping}
\end{table}

\begin{table}[]
    \newcommand{\ycbC}{1.5}
    \centering
    \fontsize{8.0}{8.0}\selectfont
    \begin{tabular}{C{\ycbC cm}|C{\ycbC cm}|C{\ycbC cm}|C{\ycbC cm}}
    \hline
    Group & from scratch & pretrained & pretrained
    +
    finetuned \\ \hline
    0     & 62.8         & \textbf{73.9}       & 70.0                 \\
    1     & 57.7         & 77.9       & \textbf{86.8}                 \\
    2     & 75           & 86.1       & \textbf{93.4}                 \\ \hline
    Mean  & 65.2         & 79.3       & \textbf{83.4}                 \\ \hline
    \end{tabular}
    \caption{Effect of ShapeNet6D for pre-training on the LineMOD dataset. The variety of shape and appearance priors improves generalizability by large margins.}
    \label{tab:linemod_cross_domain+effect_of_pretrain}
\end{table}

\subsection{Benchmark Datasets}
The LineMOD \cite{hinterstoisser2011multimodal} and the YCB-Video \cite{calli2015ycb} are two popular datasets for 6D object pose estimation. The LineMOD dataset contains 13 videos of 13 low-textured objects, while the YCB-Video dataset consists of 92 RGBD videos of 21 YCB objects. For the few-shot pose estimation problem, we select 16 shots for each object for pose estimation. We also follow the strategy of other well-established few-shot problems, i.e., segmentation, and split the dataset into different groups. Specifically, we split the objects into three groups for each dataset and select one for testing and the remaining two for training each time (see the supplementary material for details). 
\subsection{Evaluation Metrics}
The average distance metrics ADD and ADDS are widely used for performance evaluation of 6D pose estimation. For an object $\mathcal{O}$ consists of vertexes $v$, the ADD of asymmetric objects with the predicted pose $R$, $T$ and ground truth pose $R^*$, $T^*$ is calculated by:
\begin{equation}
    \label{eqn:ADD}
    \textrm{ADD} = \frac{1}{m} \sum_{v \in \mathcal{O}} || (Rv+T) - (R^*v + T^*) || .
\end{equation}
For symmetric objects, the ADDS based on the closest point distance is defined as:
\begin{equation}
    \label{eqn:ADDS}
    \textrm{ADDS} = \frac{1}{m} \sum_{v_1 \in \mathcal{O}} \min_{v_2 \in \mathcal{O}}{|| (Rv_1+T) - (R^*v_2 + T^*) ||} .
\end{equation}
In the YCB-Video dataset, the area under the accuracy-threshold curve obtained by varying the distance threshold (ADDS and ADD AUC) is reported following \cite{he2020pvn3d,FFB6D,xiang2017posecnn}. In the LineMOD datasets, we report the distance less than 10\% objects diameter recall (ADD-0.1d) as in \cite{hinterstoisser2012model,peng2019pvnet}.

\subsection{Baselines}
Possible solutions to the few-shot 6D object pose estimation problem include local image feature matching, point cloud registration, and template matching. We select the state-of-the-art solution in each direction as our baseline. 

\textbf{LoFTR} \cite{sun2021loftr} is a detector-free deep learning architecture for local image feature matching. It uses the self- and cross-attention layers in Transformers to obtain high-quality matches. 

\textbf{PREDATOR} \cite{huang2021predator} is a  neural architecture for pairwise 3D point cloud registration with deep attention to the overlap region. It learns to detect the overlap region between two unregistered scans and focus on that region when sampling feature points. 

\textbf{Template Matching.} Template matching approaches \cite{huttenlocher1993comparing,gu2010discriminative,hinterstoisser2011gradient} discrete pose estimation problem into classification problem. These approaches rely on CAD models to generate thousands of templates and retrieve the closest one to the scene. However, we eliminate the dependency of precise object CAD models in our problem. Besides, capturing, labeling, and storing thousands of support shots are time- and storage-consuming. We assign the view with rotation closest to the ground truth and the center shift as translation to reveal the upper bound of these approaches.

For a fair comparison, all baselines and the proposed one are not equipped with iterative refinement, e.g., ICP \cite{besl1992_ICP}.

\subsection{Training and Implementation}
We crop object patches with ground-truth bounding boxes for our model and resize them to $255\times255$ as input. The correspondence is optimized by negative log-likelihood loss \cite{sarlin2020superglue}. For a fair comparison, we pretrained all models on ShapeNet6D with online data augmentation for two epochs and fine-tuned on benchmark datasets for five epochs. We select 16 different views for each object as support images.

\subsection{Benchmark Results}
\textbf{Results on LineMOD and YCB-Video datasets.}
Quantitative results on the YCB-Video and the LineMOD dataset are shown in Table \ref{tab:ycb_result} and Table \ref{tab:linemod_result} respectively. Thanks to the joint reasoning of appearance and geometric relationship between the support and query images, our method outperforms the state-of-the-art local image feature matching method and point cloud registration algorithms by large margins. Some qualitative results are shown in Figure \ref{fig:Qualitative_Results}.

\textbf{Domain generalization.} As is shown in Table \ref{tab:linemod_cross_domain+effect_of_pretrain}, our model trained on ShapeNet6D with online data augmentation is $4.1\%$ behind the fine-tuned one. Considering the small shape and appearance diversity in the LineMOD dataset, compared with ShapeNet6D, we think the performance drop mainly comes from the domain gap. More future works are expected to bridge this gap to fully explore the power of shape and appearance diversity in ShapeNet6D, e.g., designing domain invariant algorithms.


\subsection{Ablation Study}

\textbf{Effect of pre-training on the large-scale ShapeNet6D.} As shown in Table \ref{tab:linemod_cross_domain+effect_of_pretrain}, FS6D-DPM trained on ShapeNet6D outperforms the one trained from scratch on the LineMOD dataset by a large margin ($+11\%$), proving the efficacy of the shape and appearance diversity resides in ShapeNet6D.


\textbf{Effect of online texture blending.} As shown in Table \ref{tab:online_wrapping}, the proposed online texture blending provides diverse texture prior and improves the performance on texture-rich objects in the YCB-Video dataset by large margins.






\section{Discussion and Limitations}
In this work, we study a challenging \textit{open-set} problem, the few-shot 6D object pose estimation. We point out the essence of appearance and geometric information to tackle the problem and propose FS6D-DPM as a solid baseline to solve it. Furthermore, we show that prior from diverse shapes and appearances are crucial to the generalizability of few-shot 6D pose estimation algorithms and introduce a large-scale dataset (ShapeNet6D) for network pre-training. An online texture blending augmentation is proposed to bridge the domain gap as well.

However, there are still some limitations in this work. Firstly, we focus on the pose estimation problem and rely on object detection algorithms to crop out the region of interested objects. Though various off-the-shelf few-shot object detection algorithms \cite{kang2019few} are available, a joint framework is more practical. Secondly, despite being diverse in shape and appearance, the proposed large-scale ShapeNet6D is synthesis, and the domain gaps problem is not tackled yet. Future directions include domain invariant pose estimation algorithms or large-scale real-world datasets. Lastly, there is still a significant performance gap between few-shot algorithms and those trained under the \textit{close-set} setting. We expect more future research, e.g., leveraging 3D keypoint-based techniques \cite{FFB6D,he2020pvn3d} to bridge this gap.

\noindent\textbf{Acknowledgements} This work is supported by Guangzhou Okay Information Technology with the project GZETDZ18EG05.

\appendix

\begin{table*}[]
    \centering
    \fontsize{8.}{8.}\selectfont
\begin{tabular}{llcccccccccccccc}
\toprule
                                     &              & ape           & bench.     & camera        & can           & cat           & driller       & duck          & \textbf{eggbox} & \textbf{glue} & holep.   & iron          & lamp          & phone         & Mean          \\\midrule
w/o ref.                 & Ours                 & 74.0                    & 86.0                          & 88.5                       & 86.0                    & \textbf{98.5}           & 81.0                        & 68.5                     & \textbf{100.0}                      & \textbf{99.5}                     & \textbf{97.0}                   & \textbf{92.5}            & 85.0                     & \textbf{99.0}             & 88.9                     \\ \midrule
\multirow{2}{*}{w/ ref.} & LatentFusion \cite{park2020latentfusion}         & \textbf{88.0}           & \textbf{92.4}                 & 74.4                       & 88.8                    & 94.5                    & 91.7                        & 68.1                     & 96.3                                & 94.9                              & 82.1                            & 74.6                     & 94.7                     & 91.5                      & 87.1                     \\
                         & Ours+ICP             & 78.0                    & 88.5                          & \textbf{91.0}              & \textbf{89.5}           & 97.5                    & \textbf{92.0}               & \textbf{75.5}            & 99.5                                & \textbf{99.5}                     & 96.0                            & 87.5                     & \textbf{97.0}            & 97.5                      & \textbf{91.5}   \\
\bottomrule
\end{tabular}
    \caption{Quantitative evaluation of different few-shot (16 shots) 6D pose estimation on the LineMOD dataset with ground truth segmentation. w/o ref.: without iterative refinement; w/ ref.: with iterative refinement. Symmetry objects are in bold.}
    \label{tab:lm_gt_seg}
\end{table*}

\begin{table*}[]
    \centering
    \fontsize{8.}{8.}\selectfont
\begin{tabular}{cccccc|ccccc|ccccc}
\hline
\multicolumn{6}{c|}{Group 0}                     & \multicolumn{5}{c|}{Group 1}                              & \multicolumn{5}{c}{Group 2}               \\ \hline
ape  & benchvise & camera & can  & cat  & mean & driller & duck & \textbf{eggbox} & \textbf{glue} & mean & holepuncher & iron & lamp & phone & mean \\ \hline
70.5 & 82.5      & 72.5   & 46.5 & 78.0 & 70.0 & 87.0    & 60.5 & 100.0           & 99.5          & 86.8 & 94.0        & 88.0 & 94.5 & 97.0  & 83.4 \\
\hline
\end{tabular}
    \caption{Detailed results of our method on the LineMOD dataset. Symmetry objects are in bold.}
    \label{tab:lm_details}
\end{table*}

\renewcommand{\arraystretch}{1.3}
\begin{table}[tp]
    \centering
    \fontsize{8.}{8.}\selectfont
\begin{tabular}{l|lllll}
\hline
\# Views & 1   & 4   & 8   & 16  & 32   \cr\hline
ADDS AUC$\uparrow$      & 79.6 & 87.3 & 87.9 & 88.4 & 88.6  \cr\hline
\end{tabular}
    \vspace{-1.1em}
    \caption{Effect of number of support views on the YCB-Video. The mean ADD-S AUC results are reported.}
    \label{tab:views}
    \vspace{-1.3em}
\end{table}

\section{Appendix}
\subsection{More Results}
\textbf{Model running time.}
We run our model on a single NVIDIA GeForce RTX 2080Ti GPU. For each support view, it takes an average time of 72 ms for neural network forwarding and 113 ms for pose alignment using the Umeyama algorithm \cite{umeyama1991least} with RANSAC \cite{fischler1981random}.

\textbf{Pose estimation with ground-truth segmentation.}
In the main paper, we utilize relaxed ground-truth object bounding boxes to crop out regions of interested objects from the query scene for pose estimation. While LatentFusion \cite{park2020latentfusion} utilizes stricter ground-truth segmentation to segment out objects, we report our results on LineMOD dataset following their setting. Specifically, We use our model trained only with ShapeNet6D without fine-tuning on the real LineMOD dataset. As is shown in Table \ref{tab:lm_gt_seg}, our model without any refinement already surpasses iterative refined LatentFusion. Equipped with post-refinement by ICP, our model obtains further improvement. Moreover, our model (0.34 fps) is 18X faster than LantentFusion (0.018 fps) on a RTX 2080Ti, when both use 16 support views.

\textbf{Effect of the different number of support views.} We ablate the effect of the different number of support views in Table \ref{tab:views}. As is shown in the table, our algorithm gets better performances when the number of support views increases. Moreover, it only gains margin performance when we have more than 16 views, which shows that our algorithm does not need too many support views and can get good pose results under the few-shot setting.

\textbf{Details results on the LineMOD dataset.} See Table \ref{tab:lm_details}.

\textbf{Visualization of ShapeNet6D}
Example images in ShapeNet6D are shown in Fig. \ref{fig:vis_shapenet6d}.

\subsection{Implementation Details}
\textbf{Grouping information of benchmark datasets.}
We split the LineMOD dataset into three groups. Objects in different groups have no intersection. During network fine-tuning, we select two groups for training and one group as novel objects for testing. The group information of the LineMOD dataset is shown in Table \ref{tab:linemod_group_info}. We split the YCB-Video dataset into three groups in a similar way. Group information of the YCB-Video dataset is shown in Table \ref{tab:ycb_group_info}.

\textbf{Support views selection.}
\label{sec:frs}
We select support views from the training set since we do not have the real-world objects in the LineMOD and YCB-Video datasets to capture the support views. We select 16 support views using the farthest rotation sampling for each object to ensure that each part of the object is visible. Specifically, we initialize the set of selected views with a random view from the training set for each object. We then add another object view with the farthest rotation distance from views in the selected set. We repeat this procedure until 16 views of the target object are obtained. We define the distance between two rotations as the Euclidean distance between two unit quaternions following \cite{ravani1983motion,huynh2009metrics}. The formula is as:
\begin{equation}
    D(q_1, q_2) = \min\{||q_1 - q_2||, ||q_1+q_2||\}.
\end{equation}
where $||\cdot||$ denotes the Euclidean norm and $q_1, q_2$ the two unit quaternions. 

Given the target object's mask labels and pose parameters in the selected support views, we crop out the object region and transform the object point cloud back to the object coordinate system to serve as a reference frame to define the 6D object pose.

\subsection{Fast Registration of Novel Objects}
Given a novel object and an RGBD sensor with known intrinsic parameters, we can quickly obtain support views of the novel object in several ways. We provide some examples as follows:

\begin{figure*}
  \centering
     \includegraphics[width=0.95\linewidth]{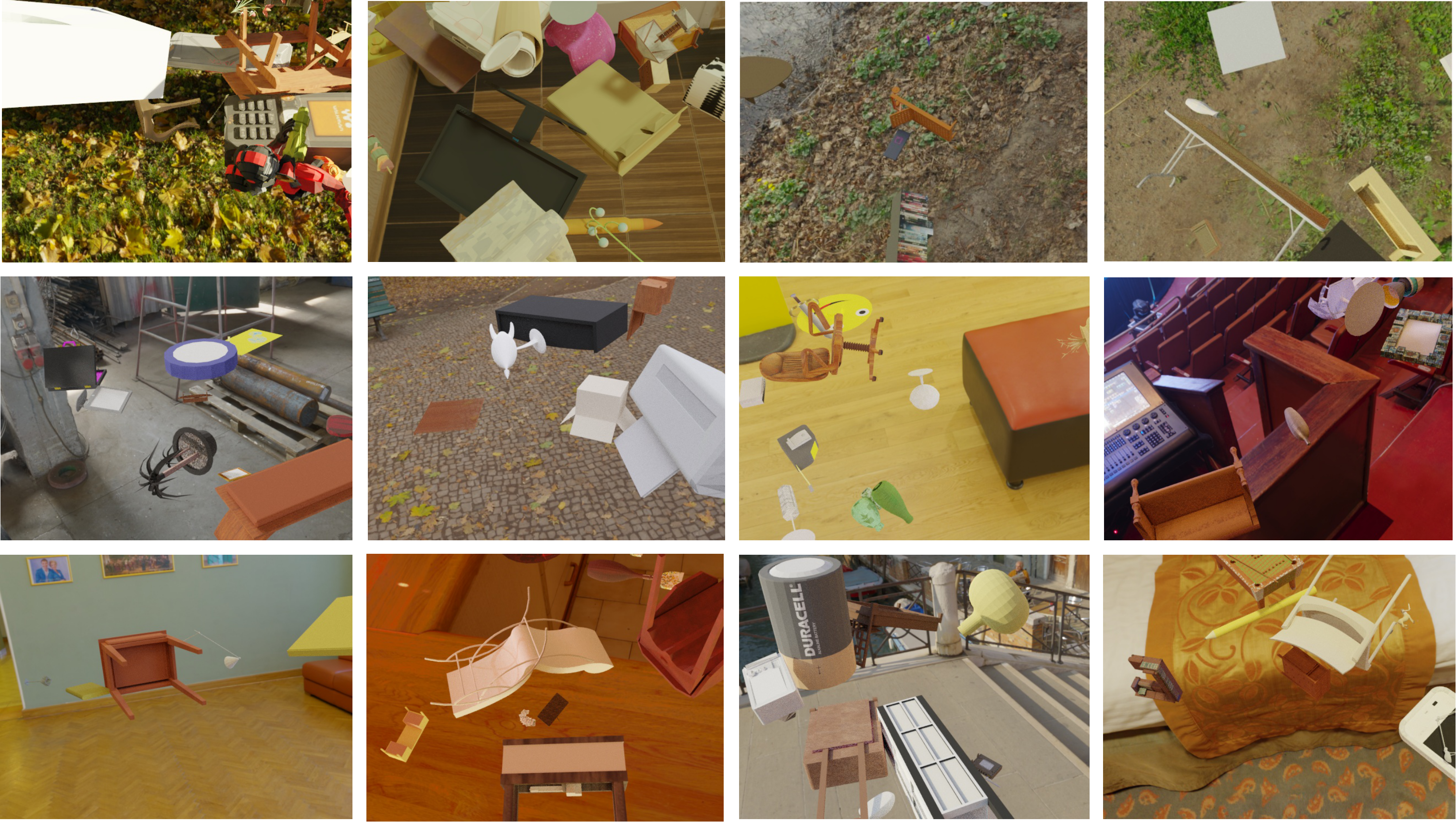}
     \caption{
        Example scene images in our ShapeNet6D dataset.
     }
     \label{fig:vis_shapenet6d}
\end{figure*}

\begin{table*}[]
    \centering
    \fontsize{8.}{8.}\selectfont
        \begin{tabular}{cc}
        \toprule
        Group  & Objects \\ \midrule
        0 & 002 master chef can, 003 cracker box, 004 sugar box, 005 tomato soup can, 006 mustard bottle, 007 tuna fish can, 008 pudding box\\
        1 & 009 gelatin box, 010 potted meat can, 011 banana, 019 pitcher base, 021 bleach cleanser, 024 bowl, 025 mug  \\
        2 & 035 power drill, 036 wood block, 037 scissors, 040 large marker, 051 large clamp, 052 extra large clamp, 061 foam brick \\
        \bottomrule
        \end{tabular}
    \caption{Group information of the YCB-Video dataset.}
    \label{tab:ycb_group_info}
\end{table*}

\begin{table}[]
    \centering
    \fontsize{8.}{8.}\selectfont
        \begin{tabular}{cc}
        \toprule
        Group  & Objects \\ \midrule
        0 & ape, benchvise, camera, can, cat \\
        1 & driller, duck, eggbox, glue      \\
        2 & holepuncher, iron, lamp, phone  \\
        \bottomrule
        \end{tabular}
    \caption{Group information of the LineMOD dataset.}
    \label{tab:linemod_group_info}
\end{table}

\textbf{Select from an RGBD video of the novel object.} The most simple way is to select support views from an RGBD video of the target object. Specifically, we first place the target object in the center of a clean plane and then capture a video by slowly moving the camera around the object. We use the first frame to define the object coordinate system. Specifically, we mask out the object region by removing the background plane with a plane detection algorithm \cite{feng2014fastplane} or least-square-fitting of a plane on the scene point cloud. We define the object coordinate system based on the object point cloud of the first frame. Then, we calculate the pose between the following frame and the first frame. Since the pose difference between adjacent frames of a video is small and the scene background is a clean plane, we can utilize registration algorithms, i.e., ICP \cite{besl1992_ICP}, Go-ICP \cite{yang2015goicp} to calculate the relative pose parameters between adjacent frames and obtain the pose parameters between each frame and the first frame. Finally, we can select support views by the farthest rotation sampling algorithm as in Section \ref{sec:frs}. 

To further improve the accuracy of relative pose parameters, we can put the object on a marker board (a plane with several markers on it) and utilize markers to obtain more accurate relative poses.

\textbf{Collect with the robot arm.} For robotic manipulation, we have a robot arm with a camera in hand. We first calibrate the robot arm and the camera between an observed region with a marker board. We define several viewing points with known pose parameters. We then place the novel target object to the observed region and utilize the robot arm to move the camera to those predefined viewing points to capture support views of the novel objects. The pose parameters of support views will be more accurate due to the robustness of the robotic manipulation system.


{\small
\bibliographystyle{ieee_fullname}

}

\end{document}